\documentclass[conference]{IEEEtran}
\IEEEoverridecommandlockouts

\usepackage{cite}
\usepackage{amsmath,amssymb,amsfonts}
\usepackage{graphicx}
\usepackage{textcomp}
\usepackage{xcolor}
\usepackage{listings}
\usepackage{url}
\def\BibTeX{{\rm B\kern-.05em{\sc i\kern-.025em b}\kern-.08em
    T\kern-.1667em\lower.7ex\hbox{E}\kern-.125emX}}
\newcommand{\sysname}{AgenticRei}
\newcommand{\hide}[1]{} % hide text

\lstdefinestyle{ttlpolicy}{
  basicstyle=\ttfamily\footnotesize\bfseries,
  breaklines=true,
  breakatwhitespace=false,
  columns=flexible,
  keepspaces=true,
  frame=single,
  framerule=0.4pt,
  backgroundcolor=\color{gray!4},
  commentstyle=\ttfamily\itshape
  morecomment=[l]{\#},
  xleftmargin=4pt,
  xrightmargin=4pt,
  aboveskip=4pt,
  belowskip=2pt,
}

\begin{document}

\title{Deontic Policies for Runtime\\Governance of Agentic AI Systems}

\author{%
\IEEEauthorblockN{Anupam Joshi}
\IEEEauthorblockA{\textit{CSEE Department}\\
UMBC\\
Baltimore, MD, USA\\
joshi@umbc.edu}
\and
\IEEEauthorblockN{Tim Finin}
\IEEEauthorblockA{\textit{Center for AI}\\
UMBC\\
Baltimore, MD, USA\\
finin@umbc.edu}
\and
\IEEEauthorblockN{Karuna Joshi}
\IEEEauthorblockA{\textit{Information Systems  Department}\\
UMBC\\
Baltimore, MD, USA\\
kjoshi1@umbc.edu}
\and
\IEEEauthorblockN{Lalana Kagal}
\IEEEauthorblockA{
\textit{CSAIL}\\
MIT\\
Cambridge, MA, USA\\
lkagal@csail.mit.edu}
}

\maketitle

% -- Abstract --------------------------------------------------------------
\begin{abstract}
Autonomous agentic AI systems driven by Large Language Models (LLMs) introduce a new class of security,
privacy, and compliance challenges: an agent that can invoke tools,
manipulate data, install software, and coordinate with peer agents
across organizational boundaries must be constrained not just by
authentication and access control, but by the full structure of
enterprise governance. This includes specifying what agents are \emph{permitted} and
\emph{prohibited} from doing, what they are \emph{obliged} to do after
certain actions (e.g., notify the CISO), under what conditions a standing obligation may be
\emph{waived}, and which rules take precedence when policies conflict.
This governance problem exceeds what current policy
engines provide. Systems such as XACML, Rego, and Cedar address only the permit/prohibit subset of this governance structure. They do not provide obligation lifecycle management, meta-policy conflict resolution, dispensations that waive obligations in specific circumstances, and ontological reasoning over domain class hierarchies commonly found in applications such as healthcare, cybersecurity, or data privacy.  
We propose \sysname{}, which realizes key governance requirements such as obligations, dispensations, policy conflict resolutions, and reasoning over policies, as well as the basic permit/prohibit constraints. We use a deontic policy language built on the Rei framework, expressed
as OWL (Web Ontology Language) and evaluated at runtime by a high-performance 
logic engine entirely outside the LLM\@.  The same pipeline governs
both tool invocations by the agent and agent-to-agent messages.  We show through examples that deontic policies capture governance constraints around security and 
privacy that mostly cannot be expressed in current
production engines. Our approach composes naturally with industry-standard frameworks like
A2AS\@.
\end{abstract}


% ==========================================================================
\section{Introduction}
% ==========================================================================

The emergence of Agentic AI systems in the form of LLM-driven planners that invoke
external tools, coordinate with peer agents, and execute multi-step workflows has introduced governance challenges that static, prompt-level guardrails cannot meet. A recent example using Agentic AI in a Security Operations Center identifies some of these challenges~\cite{banstola2026soc}. Such systems are being deployed in areas as diverse as healthcare~\cite{savage2025open} and national security~\cite{chukkapalli2025impostors}. The industry response has converged on a shared
premise. Since LLM reasoning is non-deterministic, security must be
enforced at the \emph{point of action execution} through a deterministic
policy layer independent of the LLM.  Frameworks such as Wallarm's
A2AS~\cite{neelou2025a2asagenticairuntime}, Microsoft's Agent Governance
Toolkit~\cite{ms_agent_gov_toolkit}, and Cisco's MCP policy-enforcement
gateway in Secure Access~\cite{cisco_agent_runtime} all embody this premise.

This convergence validates the \emph{where} aspect of governance: the right place to enforce policy is at the action boundary, not in the
context window of the LLM\@. A recent joint advisory from CISA, NSA, and allied national cybersecurity agencies independently reaches the same conclusion, explicitly recommending per-invocation rather than startup-time policy evaluation~\cite{fiveeyes2026agentic}. But \emph{where} to enforce is only half the problem. As we argue in Section~\ref{sec:case}, the governance requirements that regulators, standards bodies, and early production deployments need go beyond what the policy languages of these frameworks can express. Our contribution addresses \emph{the what and the how}: the expressive power of the policy language integrated with the Semantic Web, the machine-understandable layer of the Web, and fast reasoners. We discuss how this model can be integrated with industry standards like A2AS.

A2AS implements its Codified Policies (C) pillar primarily through context augmentation and the model's own reasoning, rather than through an external policy engine, consistent with its design goal of avoiding external dependencies. The A2AS authors acknowledge a resulting limitation they term \emph{security reasoning drift}: because in-context defenses and codified policies are interpreted by the model, variations in model reasoning can lead to misinterpretation or partial compliance~\cite{neelou2025a2asagenticairuntime}. Providing the LLM with information about security guardrails does not guarantee compliance.
Externalized engines used in industry (such as OPA~\cite{opa_official} and Cedar~\cite{cedar_oopsla24})
avoid the LLM-as-enforcer problem, but expose a second limitation. Their policy languages are flat allow/deny or Access Based Access Control (ABAC) rules, which can only decide whether an action is permitted or denied. 

We argue that open, multi-agent systems need four properties
that flat-list engines cannot provide, but are captured
naturally in a policy language like Rei~\cite{kagal2003policy} including its application to domains such as pervasive context-aware privacy~\cite{ChenUbiquitous2003, Patwardhan2004}, that is based on Deontic Logic and grounded in OWL/RDF semantics (Web Ontology Language~\cite{w3c_owl}). It captures the deontic notions of permissions, prohibitions, obligations, and dispensations as first-class semantic objects in ways that can then be reasoned over to make governance decisions. 

% ==========================================================================
\section{Governance Gap and Case for Deontic Policies}
\label{sec:case}
% ==========================================================================

\subsection{The Governance Gap}
\label{sec:gap}

Regulators, standards bodies, and early production deployments have
begun to articulate what governing an autonomous agent actually
requires, and the emerging consensus points well beyond access control.
Three themes recur.

\emph{Oversight must scale with autonomy, and regulators are leaving the
mechanism to institutions.}  The World Economic Forum's progressive
governance model holds that the degree of oversight applied to an agent
should be proportional to the autonomy it is granted~\cite{wef2025agents},
a position echoed by enterprise adoption surveys reporting that a large
majority of executives plan agent deployment within three years.  Yet
the regulatory instruments that would normally supply this oversight
explicitly decline to.  The April 2026 interagency model-risk guidance
(SR~26-2), jointly issued by the Federal Reserve, FDIC, and OCC,
excludes generative and agentic AI from its scope in footnote~3, while
affirming that each institution remains responsible for governing
systems the guidance does not cover~\cite{sr2602}.  NIST's Center for
AI Standards and Innovation has launched a dedicated AI Agent Standards
Initiative to address this gap~\cite{nist_caisi}, signaling that
agent-specific standards are still being written.

\emph{Authority creep and diffuse accountability are the dominant
failure modes in practice.} In production financial services,
agentic systems deployed as decision-support have their confirmation
requirements relaxed and autonomy thresholds raised incrementally,
until a system with the action authority of a high-risk deployment is
still governed as the cautious pilot it once was ---``no one made a bad
decision; no one made the decision at all''~\cite{saxena2026finance}.
When such a decision is later challenged, accountability is diffuse:
product owns the model, engineering owns the infrastructure, and
compliance owns the policy, so no team can reconstruct who authorized
the action~\cite{saxena2026finance}.  The Five Eyes advisory identifies
the technical root of this problem as non-reproducible agent decision
chains, and calls for a durable record of what an agent did and under
whose authority~\cite{fiveeyes2026agentic}.

\emph{Standards say what to control, not how to enforce it.}  Agent
certification standards such as AIUC-1~\cite{aiuc1} define auditable
control objectives---for example, control~B006, ``prevent unauthorized
AI agent actions''---and the NIST AI Risk Management
Framework~\cite{nist_ai_rmf} organizes risk management into govern,
map, measure, and manage functions.  Both specify \emph{what} an
enforcement layer must achieve, but are deliberately silent on the
runtime mechanism.  The service-infrastructure layer is left with a
substantial implementation gap: how to translate a control objective
written in a PDF into a decision made deterministically at every action
boundary.  Closing that gap is the problem addressed in \sysname{}, and we
return to the concrete mapping of standards-to-runtime in 
Section~\ref{sec:vision}.

\subsection{Structural Requirements}
\label{sec:requirements}
A basic permission or prohibition in Rei is no more complex than its Rego or Cedar equivalent and is enforced with the same default-deny guarantee. Beyond this baseline, governance requires answering a harder question: not just whether an action is permitted, but what follows from permitting it. We identify four requirements arising from this second question in enterprise agentic deployments that are structurally inexpressible in allow/deny or ABAC rule engines.

\textbf{Obligations.}  Real-world governance attaches not just permissions, but \emph{duties} to permitted actions.  A compliance
officer who reads a regulated dataset \emph{must} log the access within
60 seconds. A system that installs software on a production host
\emph{must} notify the CISO\@. These are not access-control rules; they are behavioral obligations that arise \emph{because} access was
granted.  Policy languages like Rego \cite{opa_official} and Cedar \cite{cedar_oopsla24} \hide{\cite{cedar_vgd_24}} do not have a native obligation construct: the
policy author must create a separate rule checking whether a prior
logging event occurred, and then chain it manually.  This is a workaround to the actual challenge of managing obligations.

\textbf{Principled conflict resolution.}  In multi-organizational
settings, multiple authorities issue overlapping rules.  An
organizational policy may prohibit exporting any dataset; a
project-level policy may permit members to export a specific one; a
regulatory override may permit export in response to a legal subpoena.
Flat rule lists resolve such conflicts by evaluation order or an ad hoc
priority integer field.  Neither is semantically meaningful.  Deontic
frameworks provide \emph{meta-policies} which are rules about rules. That makes
conflict resolution explicit, authoritative, and auditable.

\textbf{Semantic grounding.} ABAC rules match the syntactic attribute
values. A rule that prohibits access to ``Protected Health Information (PHI)'' must enumerate every concrete attribute value that constitutes PHI in the system and must be
updated whenever the domain model evolves. A policy engine that can
reason over a domain ontology where a ``pediatric oncology record''
is declared to be a subclass of ``health record'' which is a subclass
of ``PHI'', can express the prohibition once, at the class level, and
have it apply automatically to all current and future subclasses without
touching the policy source.

\textbf{Dynamic, cross-authority trust.}  Open agent ecosystems require
that trust in a credential not just depend on \emph{who signed it}, but that this
trust relationship itself is expressed as policy.  An override of
an export prohibition should be granted only when the agent presents a
credential whose issuer is the \emph{compliance authority named in the
policy}, and not just any credential bearing the claim ``is\_compliance\_officer
= true.''  This cross-pillar composition, in which the policy names
trusted credential issuers and the cryptographic infrastructure delivers
signed artifacts, cannot be expressed in flat ABAC rules without
hard-coding issuer identifiers outside the policy. South et al.~\cite{south2025authenticateddelegation} extend OAuth~2.0 and OpenID Connect with agent-specific delegation credentials precisely because the base protocols, while establishing who an agent is and on whose behalf it acts, do not by themselves express which authority's credential a policy treats as trusted for a given action. This is a policy-governed trust relationship that \sysname{} makes explicit.

% ==========================================================================
\section{Architecture}
% ==========================================================================

\begin{figure}[t]
\centering
\includegraphics[width=\columnwidth]{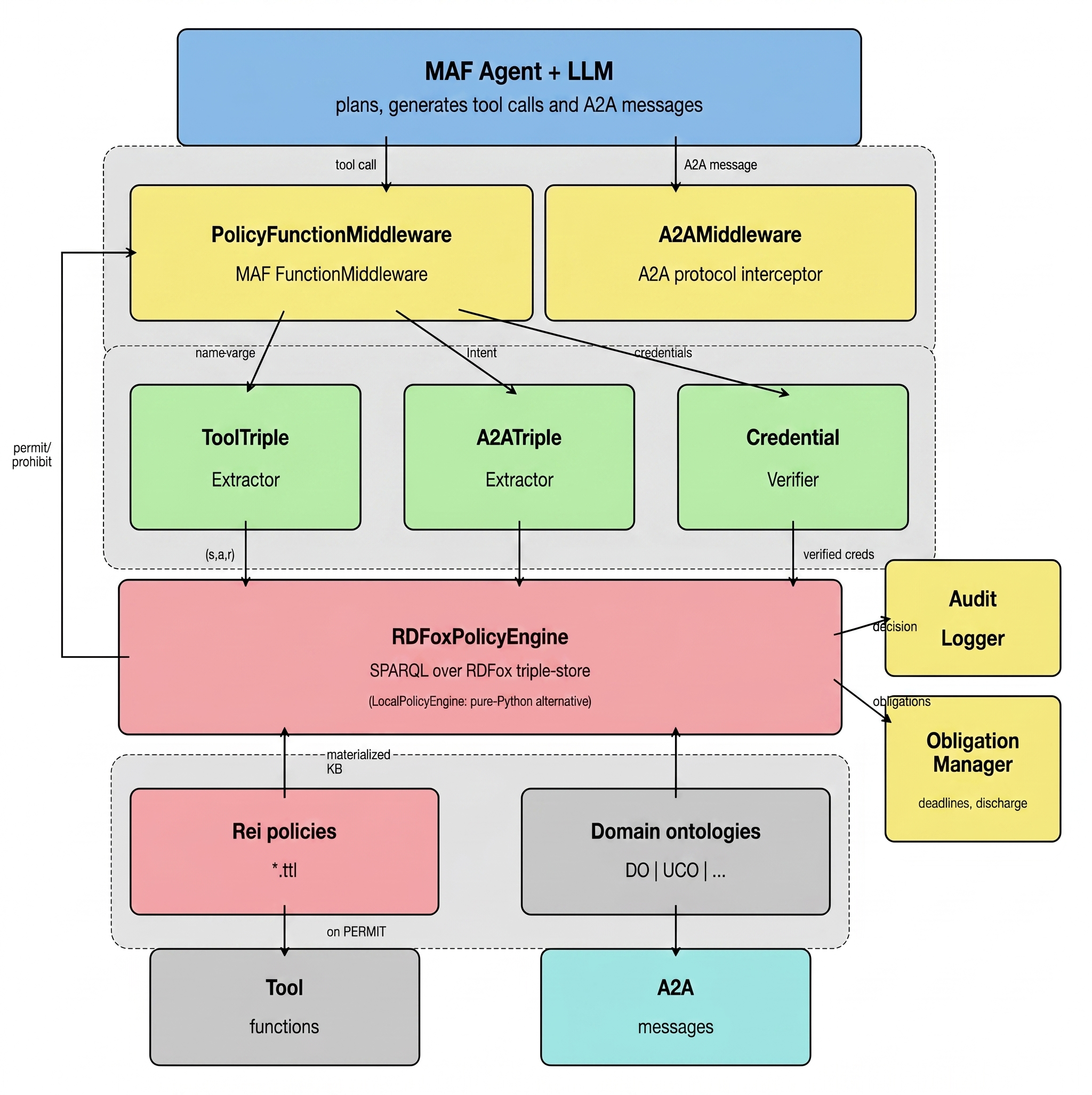}
\caption{%
  \sysname{} runtime architecture.  Both tool-call interception and
  A2A message interception share the same \texttt{PolicyEngine},
  \texttt{AuditLogger}, and \texttt{ObligationManager}.  %
}
\label{fig:arch}
\end{figure}

\sysname{} enforces deontic policies at the action boundary of any
agent framework.  The architecture, shown in Fig.~\ref{fig:arch},
follows a three-step \emph{extract--evaluate--apply} contract. We have built a prototype that demonstrates the approach end to end. It loads Rei-encoded policies and domain ontologies, intercepts an agent action and extracts it to a $\langle$subject, action, resource$\rangle$ triple, evaluates that triple against an RDFox-based policy engine, and returns a verdict with any attached obligations. The prototype implements permission, prohibition, obligation, dispensation, and meta-policy priority resolution over the Rei ontology, with subclass reasoning supplied by RDFox's OWL/RDFS entailment. Credential verification is currently simulated through trusted-issuer matching rather than cryptographic signature checking; integration with production agent runtimes (the A2A protocol and the Microsoft Agent Framework) and with cryptographic credential verification is ongoing work.

\textbf{Extract.}  A \emph{TripleExtractor} intercepts every outbound
action, whether by a tool invocation or an agent-to-agent (A2A)
message, at the framework's middleware boundary, and maps it to a
concrete $\langle\textit{subject},\,\textit{action},\,\textit{resource}\rangle$
triple.  This extraction happens after the agent framework has already
validated the call against its schema, so the extractor operates on a
typed, structured invocation, not on raw LLM text.  Credentials accompanying the request are extracted in the same step and checked against the issuers the policy trusts; those that fail the check are discarded before the engine sees them.

\textbf{Evaluate.}  A \emph{PolicyEngine} queries a triple-store loaded
with Rei-encoded deontic rules and pluggable domain ontologies. The
engine returns a verdict of \textsc{permit}, \textsc{prohibit}, or
\textsc{default-deny}, together with any obligations attached to a
permitted action via the Rei \texttt{deontic:provision} construct.
Every internal failure, such as an exception, timeout, or missing rule, produces
\textsc{default-deny}, never an exception.

\textbf{Apply.}  The middleware either allows execution to proceed (on
\textsc{permit}), appending obligation text to the action result for the
agent to act on, or short-circuits the invocation and returns a
structured policy-violation message (on \textsc{prohibit} or
\textsc{default-deny}).  The LLM has no role in this decision.

Every decision is serialized to a structured audit record capturing the matched rule, round-trip latency, the credential issuers presented, and a hash of the loaded policy Knowledge Base (KB) computed from its TTL sources. The KB hash identifies which policy version was in effect for a decision, supporting forensic queries of the form "what rules were in effect when this decision was made." The record logs attribute names rather than their values, and is an append-only log rather than a tamper-evident ledger; full decision reproducibility would additionally require logging the staged request graph.
 Since the policy KB contains stable governance rules
(not per-agent session state), the description-logic engine can
materialize conclusions at load time and reduce each per-action query to
a triple-pattern lookup, yielding sub-millisecond decision latency. In preliminary measurements on a single host (RHEL 9, RDFox 7.5, the policy engine reached over localhost HTTP), each prohibition/permission query pair plus any obligation lookup completes end to end in under 10 ms, with the RDFox query execution itself under a millisecond and the remainder being HTTP round-trip overhead. This is well within the bounds required for the interception of synchronous actions in production environments.

% ==========================================================================
\section{Deontic Policies in Practice}
% ==========================================================================
\label{sec:scenarios}
The policies in \sysname{} are expressed in the Rei deontic
framework~\cite{kagal2003policy, ChenUbiquitous2003}.  Rei
provides four modalities directly relevant to enterprise agentic
governance: \emph{permissions}, \emph{prohibitions}, \emph{obligations}
(attached to a permission via \texttt{deontic:provision}), and
\emph{dispensations} (releases an agent from an obligation).
Meta-policies are first-class constructs: a
\texttt{metapolicy:RulePriority} triple gives a project-level permission
precedence over an organization-wide prohibition when both fire on the same
request, replacing the evaluation-order heuristics of flat rule engines.

Domain semantics are supplied through pluggable ontology fragments
loaded alongside the Rei rules.  A prohibition expressed over a class
(e.g.,\ \texttt{phi:PHI}) fires automatically for any resource type
declared as a subclass in the loaded domain-ontology
fragment. The policy author does not need to enumerate subclasses.
Adding a new ontology fragment extends the coverage of existing rules
without touching the policy source.

We present the policy examples below to expose a governance property that is 
inexpressible in flat-list engines.

\subsection*{P1 --- Basic Permission and Prohibition}

The simplest case: any agent may read a public document; no agent may
delete a user record.  In Rei, permissions and prohibitions are specified using \texttt{deontic:Permission} and \texttt{deontic:Prohibition}. The rules below are instances of them and use concepts defined by the domain ontology. This can also be expressed in Rego or Cedar; P1 establishes the baseline.
The prefix bindings shown below are used throughout the examples in this
section; subsequent listings omit them for brevity.

\bigskip
\begin{lstlisting}[style=ttlpolicy]
# Rei ontology namespaces (used in all examples).
@prefix deontic <http://cs.umbc.edu/~lkagal1/rei/ontologies/ReiDeontic.owl#>.
@prefix constraint:<http://cs.umbc.edu/~lkagal1/rei/ontologies/ReiConstraint.owl#>.
@prefix metapolicy:<http://cs.umbc.edu/~lkagal1/rei/ontologies/ReiMetaPolicy.owl#>.
@prefix policy <http://cs.umbc.edu/~lkagal1/rei/ontologies/ReiPolicy.owl#>.
@prefix entity <http://cs.umbc.edu/~lkagal1/rei/ttlp/ReiEntity.owl#>.
@prefix action: <http://cs.umbc.edu/~lkagal1/rei/ontologies/ReiAction.owl#>.
@prefix rdf: <http://w3.org/1999/02/22-rdf-syntax-ns#>.
@prefix rdfs: <http://w3.org/2000/01/rdf-schema#>.
@prefix demo: <http://example.org/policy-demo#>.

# Domain actions are typed as Rei domain actions.
demo:ReadAction   a action:DomainAction .
demo:DeleteAction a action:DomainAction .

# variable
demo:AnyActor a entity:Variable .

demo:Perm_ReadPublicDoc a deontic:Permission ;
    deontic:actor      demo:AnyActor ;
    deontic:action     demo:ReadAction ;
    deontic:constraint demo:IsPublicDocument .

demo:Proh_DeleteUser a deontic:Prohibition ;
    deontic:actor      demo:AnyActor ;
    deontic:action     demo:DeleteAction ;
    deontic:constraint demo:IsUserRecord .

demo:BasicPolicy a policy:Policy ;
    policy:grants  demo:Perm_ReadPublicDoc ;
    policy:grants  demo:Proh_DeleteUser ;
    policy:defaultBehavior metapolicy:ExplicitPermImplicitProh .
\end{lstlisting}
\bigskip

The \texttt{policy:defaultBehavior} \texttt{metapolicy: ExplicitPermImplicitProh}
implements the default-deny posture: anything not explicitly permitted is
denied.

\subsection*{P2 --- Obligation via \texttt{deontic:provision}}

An agent may install software on a managed host, but doing so is
permitted only if a CISO-notification obligation is accepted. The
obligation is attached to the permission via \texttt{deontic:provision};
it fires \emph{because} the action was permitted, not as a separate
check. We are not aware of any production enforcement engine that provides this. Rego and Cedar have no
obligation construct, and while the W3C's Open Digital Rights Language (ODRL)~\cite{odrl22} defines a duty in RDF, it specifies no enforcement model or lifecycle management.

\bigskip
\begin{lstlisting}[style=ttlpolicy]
demo:NotifyAction          a action:DomainAction .
demo:InstallSoftwareAction a action:DomainAction .

demo:Ob_NotifyCISO a deontic:Obligation ;
    deontic:actor   demo:AnyActor ;
    deontic:action  demo:NotifyAction ;
    deontic:obligedTo demo:CISORole ;
    rdfs:comment "Notify CISO of software installation" .

demo:Perm_InstallSoftware a deontic:Permission ;
    deontic:actor    demo:AnyActor ;
    deontic:action   demo:InstallSoftwareAction ;
    deontic:constraint demo:IsManagedHost ;
    deontic:provision  demo:Ob_NotifyCISO .

demo:SoftwarePolicy a policy:Policy ;
    policy:grants          demo:Perm_InstallSoftware ;
    policy:defaultBehavior metapolicy:ExplicitPermImplicitProh .
\end{lstlisting}
\bigskip

The middleware intercepts the install action, evaluates the policy,
permits execution, and surfaces the \texttt{Ob\_NotifyCISO} obligation
to the agent as part of the tool result.  An \texttt{ObligationManager}
registers the obligation with a deadline and fires an escalation
callback if it is not discharged.

\subsection*{P3 --- Policy conflicts, Meta-Policy, \hide{and} Cross-Pillar Composition}

This example demonstrates the full power of the deontic framework and its composition with the A2AS B pillar.
An organizational policy prohibits exporting any regulated dataset.  An agent that presents a compliance-issued  credential that is signed by the issuer \emph{named in the policy itself} is permitted to export, because a \texttt{metapolicy:RulePriority} gives the credential-gated permission precedence over the organization-wide prohibition.  An agent without such a credential, or with a credential from an untrusted issuer, is denied.

\bigskip
\begin{lstlisting}[style=ttlpolicy]
demo:ExportAction a action:DomainAction .

# Organization-wide prohibition
demo:Proh_ExportPII a deontic:Prohibition ;
    deontic:actor      demo:AnyActor ;
    deontic:action     demo:ExportAction ;
    deontic:constraint demo:IsRegulatedDataset .

# Trusted issuer named in the policy (B-pillar)
demo:TrustedComplianceCA a entity:Agent ;
    demo:issuerDID "did:web:acme-compliance.example.org" .

# Per-request credential fact: agent presents a credential from the named CA 
demo:HasComplianceWaiver a constraint:SimpleConstraint ;
    constraint:subject   demo:AnyActor ;
    constraint:predicate demo:credentialIssuedBy ;
    constraint:object    demo:TrustedComplianceCA .

# Higher-priority permission: credential-bearing agents may export
demo:Perm_ExportWithApproval a deontic:Permission;
    deontic:actor      demo:AnyActor ;
    deontic:action     demo:ExportAction ;
    deontic:constraint demo:HasComplianceWaiver,    
                       demo:IsRegulatedDataset .

# Meta-policy: the approved-export permission overrides the prohibition
demo:PriorityApprovalOverProh a metapolicy:RulePriority ;
    metapolicy:greaterPriority demo:Perm_ExportWithApproval ;
    metapolicy:lesserPriority demo:Proh_ExportPII.

demo:ExportPolicy a policy:Policy ;
    policy:grants          demo:Perm_ExportWithApproval ;
    policy:grants          demo:Proh_ExportPII ;
    policy:rulePriority    demo:PriorityApprovalOverProh ;
    policy:defaultBehavior metapolicy:ExplicitPermImplicitProh.
\end{lstlisting}
\bigskip

This policy is inexpressible in a flat-list engine for two reasons.
First, the conflict between the prohibition and the permission is
resolved by a meta-policy, not by evaluation order or an integer field:
the \texttt{metapolicy:RulePriority} is a specified semantic object
that can itself be governed by a higher-level authority.  Second, the
trusted issuer IRI is embedded in the policy itself, so the policy names
whose signatures count, rather than trusting any attribute-bearing
credential. This is the cross-pillar composition point: the B pillar
delivers the signed credential; the C pillar decides whether that credential is from a trusted source.

\subsection*{P4 --- Ontological Reasoning over Domain Class Hierarchies}

The three preceding examples share a property with ABAC and flat-list
engines: every constraint is expressed over a specific, named resource.
In large enterprise environments, the set of regulated resource types
evolves continuously, creating new entities like dataset categories, asset
classifications, and record types. A policy author must enumerate each concrete type by hand, producing rules that
cannot handle such evolution well. This example shows how
Rei's use of the Semantic Web languages can eliminate this challenge by reasoning over type hierarchies given in domain ontologies.

A compliance policy prohibits any agent from reading Protected Health 
Information (PHI).  In \sysname{}, this prohibition is expressed once, at the
class level, against a healthcare-compliance domain ontology using the
top-level class \texttt{phi:PHI}---an illustrative fragment standing in
for a production ontology such as those built on HL7 FHIR or
SNOMED~CT\@. No enumeration of subtypes is needed in the policy.

\pagebreak 
%\bigskip
\begin{lstlisting}[style=ttlpolicy]
# Domain ontology fragment (phi_fragment.ttl), not part of the policy itself.

@prefix phi: <http://example.org/health-compliance#>.

phi:PHI                  a rdfs:Class .
phi:ClinicalRecord       rdfs:subClassOf phi:PHI .
phi:PatientTreatmentPlan rdfs:subClassOf phi:ClinicalRecord .
phi:GeneticTestResult    rdfs:subClassOf phi:PHI .
phi:PsychiatricEvaluation rdfs:subClassOf phi:ClinicalRecord .
\end{lstlisting}
\bigskip

The Rei policy is expressed once, against the top-level class:

\bigskip
\begin{lstlisting}[style=ttlpolicy]
demo:Var1 a entity:Variable .
demo:AnyActor a entity:Variable .

# Constraint matches any resource whose RDF type is phi:PHI or any declared subclass of phi:PHI*; resolved by the reasoner.
demo:IsPHIResource a constraint:SimpleConstraint ;
    constraint:subject   demo:Var1 ;
    constraint:predicate rdf:type ;
    constraint:object    phi:PHI .

demo:Proh_ReadPHI a deontic:Prohibition ;
    deontic:actor      demo:AnyActor ;
    deontic:action     demo:ReadAction ;
    deontic:constraint demo:IsPHIResource .

demo:PHIPolicy a policy:Policy ;
    policy:grants          demo:Proh_ReadPHI ;
    policy:defaultBehavior metapolicy:ExplicitPermImplicitProh .
\end{lstlisting}
\bigskip

As the RDFox knowledge graph system \cite{nenov2015RDFox} materializes the OWL subclass closure at load time, a
request to read a resource typed \texttt{phi:PatientTreatmentPlan}
matches \texttt{demo:IsPHIResource} without any modification to the
policy, as the reasoner has already inferred that
\texttt{phi:PatientTreatmentPlan} is a \texttt{phi:PHI}.  The same
prohibition fires for, say \texttt{phi:GeneticTestResult} and any future subclass added to the domain ontology without requiring policy modification.

In a Rego or Cedar rule set, the policy author must write an explicit
condition for each concrete resource type, or maintain a separately
managed allowlist that must be kept in sync with the domain model.
Neither mechanism supports the open-world assumption: a resource type
not in the list at authoring time is silently unprotected until a
human notices and updates the rule. In \sysname{}, extending the
domain ontology \emph{is} the policy update.

\subsection*{P5 --- Composing Prohibition, Dispensation, and Obligation: A Financial Services Example}

The preceding examples each isolate one governance property.  This
example shows them composing in a single deployment drawn from
production financial services, where agentic systems already execute
transactions autonomously~\cite{saxena2026finance}.  It directly
targets the \emph{authority creep} failure mode discussed in
Section~\ref{sec:gap}: \sysname{} makes the autonomy threshold a
governed object itself.  A payments agent is prohibited from
autonomously executing a \emph{high-value} transaction; a
treasury-officer approval credential from the authority named in the
policy provides a dispensation, and any permitted transaction that
exceeds the regulatory reporting threshold carries a mandatory
Currency Transaction Report (CTR) obligation under the Bank Secrecy
Act. Raising the autonomy threshold is now a policy change, an action that is
governed and subject to audit, rather than a silent configuration
relaxation. The \texttt{TripleExtractor} stages the transaction type
as an \texttt{rdf:type} assertion in the request graph, following the
same pattern as P4.

As with the \texttt{phi:} fragment in P4, the \texttt{fin:} ontology
below is constructed for illustration and kept deliberately small.
Mature, production-grade ontologies already capture these concepts:
the Financial Industry Business Ontology (FIBO)~\cite{fibo} models
financial transactions, monetary amounts, and parties, and the
Financial Regulation Ontology (FinRegOnt)~\cite{finregont} aligns FIBO
with legal-knowledge models and a complete OWL representation of the
XBRL regulatory-filing schema~\cite{finregont_xbrl} to express US
banking law and compliance filings. A production policy would be
grounded in and aligned with these standards; we use a self-contained
fragment here both for brevity and because the agent-action and
autonomy-threshold concepts the policy governs are not native to them.

%\pagebreak
\bigskip
\begin{lstlisting}[style=ttlpolicy]
@prefix fin: <http://example.org/finance#> .

# TripleExtractor stages rdf:type before evaluation (same pattern as P4). fin:HighValueTransaction is a subclass of fin:Transaction in the domain ontology. Domain actions and agent are typed there.
fin:ExecutePaymentAction a action:DomainAction .
fin:FileReportAction     a action:DomainAction .
fin:PaymentsAgent        a entity:Agent .

fin:IsHighValue a constraint:SimpleConstraint ;
    constraint:subject   demo:Var1 ;
    constraint:predicate rdf:type ;
    constraint:object    fin:HighValueTransaction.

# Constraint: agent presents approval from named authority.
fin:TreasuryAuthority a entity:Agent ; 
    fin:issuerDID "did:web:treasury-authority.example.org" .
fin:HasTreasuryApproval a constraint:SimpleConstraint ;
    constraint:subject   demo:AnyActor ;
    constraint:predicate demo:credentialIssuedBy ;
    constraint:object    fin:TreasuryAuthority .

# Prohibition: no autonomous high-value execution.
fin:Proh_AutoHighValue a deontic:Prohibition ;
    deontic:actor      fin:PaymentsAgent ;
    deontic:action     fin:ExecutePaymentAction ;
    deontic:constraint fin:IsHighValue .

# Permission overrides prohibition when treasury approval is present; CTR obligation fires BECAUSE the payment is permitted.
fin:Perm_ApprovedHighValue a deontic:Permission ;
    deontic:actor        fin:PaymentsAgent ;
    deontic:action       fin:ExecutePaymentAction;
    deontic:constraint   fin:HasTreasuryApproval ;
    deontic:provision    fin:Ob_FileCTR .

# Mandatory reporting obligation (Bank Secrecy Act CTR).
fin:Ob_FileCTR a deontic:Obligation ;
    deontic:actor     fin:PaymentsAgent ;
    deontic:action    fin:FileReportAction ;
    deontic:obligedTo fin:FinCEN ;
    rdfs:comment      "File CTR with FinCEN within 15 days." .

# Dispensation: a BSA "exempt person" counterparty (31 CFR 1020.315)
# WAIVES the CTR filing obligation -- the agent is no longer obliged to file. This is an obligation waiver, not a prohibition override. fin:ExemptPerson is a counterparty class in the domain ontology.
fin:IsExemptCounterparty a constraint:SimpleConstraint ;
    constraint:subject   demo:Var1 ;
    constraint:predicate rdf:type ;
    constraint:object    fin:ExemptPerson .

fin:Disp_ExemptCTR a deontic:Dispensation ;
    deontic:actor      fin:PaymentsAgent ;
    deontic:action fin:FileReportAction ;
    deontic:constraint fin:IsExemptCounterparty .

# Meta-policy 1: approved-payment permission overrides the prohibition.
fin:Priority_ApprovalOverProh a metapolicy:RulePriority ;
    metapolicy:greaterPriority fin:Perm_ApprovedHighValue ;
    metapolicy:lesserPriority  fin:Proh_AutoHighValue .

# Meta-policy 2: an exempt counterparty waives the CTR obligation.
fin:Priority_ExemptOverCTR a metapolicy:RulePriority ;
    metapolicy:greaterPriority fin:Disp_ExemptCTR;
    metapolicy:lesserPriority  fin:Ob_FileCTR .

fin:PaymentsPolicy a policy:Policy ;
    policy:grants          fin:Perm_ApprovedHighValue ;
    policy:grants          fin:Proh_AutoHighValue;
    policy:grants          fin:Ob_FileCTR ;
    policy:grants          fin:Disp_ExemptCTR ;
    policy:rulePriority    fin:Priority_ApprovalOverProh ;
    policy:rulePriority    fin:Priority_ExemptOverCTR ;
    policy:defaultBehavior metapolicy:ExplicitPermImplicitProh.
\end{lstlisting}
\bigskip

This single policy exercises every property from
Section~\ref{sec:requirements}. The prohibition with its governed
threshold addresses authority creep directly: raising the threshold is
a policy change requiring re-authorization, not a silent configuration
relaxation. The trusted issuer IRI is embedded in the policy itself
(B-pillar composition), so the policy names whose approval counts.
The CTR obligation fires \emph{because} the transaction was permitted;
a flat-list engine has no construct for a duty arising from a permitted
action. Finally, the \texttt{metapolicy:RulePriority} is a named,
auditable relationship between the two rules---the evidentiary trail
that resolves the diffuse-accountability problem of
Section~\ref{sec:gap} by recording exactly which rule from which
policy version authorized a challenged decision. The CTR obligation
also connects to the obligation-lifecycle discussion in
Section~\ref{sec:vision}: a filing receipt from FinCEN, ingested as a
W3C Verifiable Credential into the same triple-store as the obligation,
converts a statutory deadline from a tracked promise to a
machine-verifiable discharge record.

% ==========================================================================
\section{Threat Model and Scope}
% ==========================================================================

\sysname{} defends against \emph{policy-violating} actions that an
agent, whether through compromised reasoning, adversarial prompt
injection, or deliberate misuse, attempts. This includes executing a tool call or
sending an A2A message that violates governance rules. Recent overviews of attacks \cite{puppala2026agentfence,deng2025aiagentsunderthreat} motivate the need for such strict boundary interception. Enforcement is deterministic and outside the agent. The system either invokes or
permanently withholds the action, independent of what the LLM does.  A
prompt-injection attack that successfully causes the agent to
\emph{attempt} a prohibited action still produces a policy-violation
response rather than execution.

We do not defend the agent's \emph{reasoning process} from adversarial
manipulation.  Benchmarks such as AgentDojo~\cite{agentdojo},
InjecAgent~\cite{injecagent}, and ASB~\cite{asb} evaluate the resilience of
LLM reasoning against prompt injection. This is a different property, at a
different layer. The relationship is analogous to mandatory access
control in an operating system: kernel policy does not prevent a user
from being socially engineered to run a malicious program, but it
constrains what that program can do once it runs. The two layers are
complementary; neither subsumes the other. We also note that deception-based defenses \cite{cheat2025} and mandatory policy enforcement are complementary. One targets malicious external actors, the other governs authorized internal agents. 

Obligation \emph{discharge}, the agent's subsequent behavioral act of
fulfilling an issued obligation, is in an intermediate area. Obligation \emph{issuance} is fully deterministic:
when a Rei \texttt{deontic:provision} fires, the obligation is
registered before the action result is returned, independent of the
agent.  Discharge is a behavioral property monitored via deadline
tracking and post-facto audit, which are the standard governance mechanisms for
obligations at organizational and legal scales. In ongoing work, we are extending the system so that obligation enforcement is automated to the extent possible and auditing is transparent. 

% ==========================================================================
\section{Related Work}
% ==========================================================================

The Rei framework~\cite{kagal2003policy, kagal2006security} pioneered
OWL/RDF-based machine-interpretable policy reasoning and meta-policies
for conflict resolution. Recent academic work extends this direction:
SEAgent~\cite{seagent_mac} applies Mandatory Access Control to
agent-tool interactions via information-flow graphs; ShieldAgent~\cite{shield_agent}
extracts verifiable rules from policy documents, represents them as
linear temporal logic (LTL) constraints, and organizes them into
action-based probabilistic rule circuits over which it runs
probabilistic inference to verify an agent's action trajectory; and
SAGA~\cite{syros2026saga} provides user-controlled agent lifecycle
management with a cryptographic mechanism for deriving access-control
tokens that govern agent-to-agent interaction. Veriguard~\cite{miculicich2025veriguard} pairs offline synthesis and formal verification of a behavioral policy with online runtime monitoring of each proposed agent action against that pre-verified policy.

Our work differs from SEAgent by replacing
ABAC attribute matching with ontology-grounded deontic reasoning, and
from trajectory-based approaches by providing per-action deterministic
enforcement rather than global trajectory verification. Most directly comparable in recent literature are Progent~\cite{progent2025},
which enforces least-privilege tool-call control via a JSON DSL
deterministically outside the LLM, AgentSpec~\cite{agentspec2025},
a lightweight trigger-predicate DSL for runtime safety enforcement across
code, embodied, and autonomous-vehicle agents, and PCAS~\cite{palumbo2026pcas} which uses a Datalog based approach.  They share our
\hide{architectural} premise of deterministic per-action enforcement, but do not
provide deontic obligations, meta-policy conflict resolution, or
ontological reasoning.  MI9~\cite{wang2025mi9}, developed in a
financial-services context, is a broader infrastructure-agnostic
runtime governance framework that combines agent-semantic telemetry,
continuous authorization, finite-state-machine conformance checking,
drift detection, and graduated containment; its Agency-Risk Index
operationalizes the principle of Section~\ref{sec:gap} that oversight
should scale with autonomy. Its conformance engine matches temporal
event patterns rather than evaluating deontic rules, so like the
trajectory-based approaches, it expresses neither obligations and
dispensations nor ontology-grounded, meta-policy conflict resolution.
The two are complementary, as MI9's telemetry and drift detection
could supply the behavioral signals that trigger \sysname{} policy
evaluation.

On the industry side, the A2AS BASIC model~\cite{neelou2025a2asagenticairuntime}
is the most directly comparable framework. Our work is an alternative realization of its Codified Policies pillar that externalizes policy
evaluation while consuming the Behavior pillar's cryptographic constructs.
OPA~\cite{opa_official} and Cedar~\cite{cedar_oopsla24} also provide
 externalized policy engines but lack obligations,
dispensations, meta-policies, and ontological reasoning. The W3C Open Digital Rights Language~\cite{odrl22} is structurally
closer to Rei, but lacks
a runtime enforcement architecture, obligation lifecycle management, and
meta-policy conflict resolution.  The
CoSAI/OASIS secure design patterns~\cite{cosai_ws4_mcp} explicitly
recommend OPA, Cedar, and OpenFGA as policy languages; 
\sysname{} is a richer, more expressive alternative in that space.   

PONDER~\cite{ponder2001} established a principle we follow: policy is
specified declaratively and independently of the mechanism that
enforces it, so policy can change without changing the underlying
implementation, with reusable role- and relationship-based structures
for large organizations, although PONDER's focus is object-oriented
systems. The complementary separation of a stable policy decision
point from the per-request enforcement point is XACML's PEP/PDP
pattern, on which \sysname{} builds~\cite{xacml3}, extended by Bertino
et al.\ for dynamic context and derived-data
control~\cite{bertino2015xacml,bertino2018datafusion}, but addresses
three gaps. First, XACML obligations are PEP-directed annotations without
lifecycle management rather than deontic duties derived from
permissions; second, conflict resolution uses per-PolicySet combining
algorithms rather than named, governable meta-policies; and finally, attribute
matching cannot be grounded in OWL class hierarchies.
OPA~\cite{opa_official} resolves conflicts through bundle-load ordering
external to the policy language; Cedar~\cite{cedar_oopsla24} through a
fixed \texttt{forbid}-overrides-\texttt{permit} default; neither
supports named, directed priority relationships between specific rules
that are first-class policy objects governable by a higher authority, as
Rei provides.

The Belnap policy logic program~\cite{hankin2009belnap,bruns2011belnap}
also rejects binary permit/deny and provides formal static
conflict-detection analysis. The approaches are
complementary: Belnap logic for pre-deployment compositional analysis, \sysname{} for runtime deontic governance.
AOPL-P~\cite{tummala2026autonomous} extends an authorization-and-obligation
policy language with penalties and answer-set-programming-based
planning, letting agents weigh the penalty of non-compliance against
goal utility and, in high-stakes situations, deliberately select a
policy-violating plan.

The governance standards that motivate this work---AIUC-1~\cite{aiuc1},
the NIST AI Risk Management Framework~\cite{nist_ai_rmf}, and the NIST
Generative AI Profile~\cite{nist_ai_600_1}---are discussed in
Section~\ref{sec:gap}; they define control objectives but are
mechanism-agnostic, and \sysname{} provides a runtime realization of
their action-governing subset (Section~\ref{sec:vision}).  Two further
bodies of work are complementary rather than competing. Threat
taxonomies such as MITRE~ATLAS~\cite{mitre_atlas} and the IBM AI Risk
Atlas~\cite{ibm_risk_atlas} catalog adversarial techniques and risk
categories for AI systems; \sysname{} constrains the impact stage of
an ATLAS-style attack chain regardless of whether an upstream technique
such as prompt injection succeeds, complementing the detection-oriented
defenses these taxonomies motivate.  Frontier model safety policies
such as Anthropic's Responsible Scaling Policy~\cite{anthropic_rsp}
operate at the training-time, capability-threshold level, and address
catastrophic model-level risk; they govern \emph{which models may be
deployed}, a concern orthogonal to \sysname{}'s governance of
\emph{what actions a deployed agent may take}.

% ==========================================================================
\section{Vision and Open Challenges}
\label{sec:vision}
% ==========================================================================

Earlier in the paper (Section~\ref{sec:gap}), we identified the governance gaps facing agentic AI systems and our current attempts to use Deontic Logic-based Security Governance frameworks in closing them. The examples in Section~\ref{sec:scenarios} show what a deontic
language can express today. We next outline what remains. Closing the
standards-to-runtime gap programmatically, from a control objective
written in a PDF to a decision made deterministically at every action
boundary, is the central challenge \sysname{} is designed to address,
and one that Koch~\cite{koch2026governance} identifies as an open
frontier.

\textbf{Federated policy delegation.}  In open agent ecosystems where
new agent types are deployed continuously, policy bases must evolve
without requiring the full rule set to be redeployed. Rei's OWL
representation makes incremental policy update tractable: adding a new
ontology fragment or a new Rei rule at runtime changes only the
relevant portion of the materialized KB, not the full closure.  The
harder open problem is a formal protocol for \emph{delegated} policy
update, in which a sub-authority can extend but not override a parent
authority's rules. This is the structural fix for the \emph{authority
creep} failure mode of Section~\ref{sec:gap}. When the autonomy
threshold is a governed object in the policy KB (as in P5), raising it
becomes a policy update that a delegated-update protocol can require a
parent authority to authorize, rather than a silent configuration
change that no one is accountable for.

\textbf{Realizing governance standards at runtime.}
Section~\ref{sec:gap} observed that agent standards specify what to
control but not how to enforce it.  \sysname{} provides a deterministic
enforcement substrate for the subset of controls expressible as
deontic constraints over agent actions. As shown in \cite{finin2008r,sharma2016representing}, these can capture security constraints important in the industry, such as mandatory access control, role based access control, and attribute based access control. For instance, AIUC-1's B006 (prevent unauthorized
agent actions) maps to Rei prohibitions under default-deny. Similarly, D
(restrict unsafe tool calls) maps to prohibitions over tool-invocation
action classes, and A (limit data access, prevent cross-tenant exposure)
maps to resource constraints with ontological grounding. Another instance is C (flag
high-risk actions for human review) mapping to the credential-gated override pattern
as in P3 and P5, where human-issued credentials gate permission. Finally, E
(log AI activity, assign accountability) maps to the structured,
policy-versioned audit record. Demonstrating
conformance to the action-governing subset of AIUC-1 or the NIST AI
RMF MANAGE function is achieved in part by deploying the corresponding
Rei policies, with the audit record serving as machine-verifiable
evidence for an auditor. This operationalizes the
standards-to-runtime translation that Koch~\cite{koch2026governance} frames
as an open challenge. We pair the enforcement layer (Rei policies) with
the evidence layer (Verifiable Credential discharge records, discussed next).

\textbf{Obligation lifecycle governance.}  The current framework issues
obligations deterministically and tracks their deadlines. When the obligation is explicitly given a dispensation, that can be handled as well. However, the
question of what \emph{evidence} constitutes discharge and how that
evidence is verified requires a richer model. Linking obligations to W3C Verifiable Presentations~\cite{w3c_vc} closes this loop: the CISO system issues a VC attesting that the notification was received; the VC is ingested into the same triple-store as the obligation; and a single SPARQL query can then answer ``show me every obligation issued this month, its deadline, and its discharge credential.''  This also works on agents cooperating across organizational boundaries, as long as there is a common root of cryptographic trust. This transforms obligation tracking from a best-effort audit trail into a machine-verifiable governance record, directly addressing the non-reproducible decision chains identified in Section~\ref{sec:gap}.  The CTR obligation in P5 illustrates the pattern concretely: a filing receipt from FinCEN, ingested as a VC, converts a statutory deadline into an auditable governance record. We are incorporating this into ongoing work.

\textbf{Policy authoring at scale.} Expressive policies are only useful if practitioners can author and maintain them without a specialist in both governance and formal logic on every team. Policy engineering is a challenge that requires experts in both governance and logic. Translating natural-language governance requirements into policies expressed in Rei, perhaps using an LLM, with formal
verification of the output, would lower the barrier to adoption.  Checking to make sure that the policies are good\cite{miculicich2025veriguard} and conflict free (e.g., ~\cite{bertino2018challenge})
provides the static analysis complement that Rei currently lacks. Recent work such as Policy-as-Prompt~\cite{kholkar2025policyprompt} demonstrates the authoring half of this pipeline, automatically extracting a structured, source-linked policy tree from unstructured design artifacts such as product and design documents; it compiles the result, however, into prompt-based LLM classifiers acting as runtime judges, reintroducing the very non-determinism that externalized engines are designed to avoid.  \sysname{} can consume such an extraction front-end while preserving deterministic, LLM-independent enforcement and the obligation and meta-policy constructs that prompt classifiers cannot represent. The World Economic Forum's \emph{agent card} concept, a structured pre-deployment documentation of an agent's authorized capabilities and governance constraints~\cite{wef2025agents}, is a natural upstream artifact from which Rei policy templates could be derived.  Combining tools for policy authoring and quality control with Rei-based runtime governance is an important direction for our future work.
% ==========================================================================
\section{Conclusion}
% ==========================================================================

We have argued that the full governance requirements of open
Agentic AI systems relating to security, privacy, and compliance are beyond
what current policy engines can express. We identified some needed constructs and argued that a
deontic policy language like Rei provides them.  \sysname{} demonstrates that such policies can
be evaluated at the execution time of the action, entirely outside the LLM, with
low-millisecond latency. Preliminary measurements show sub 10-ms end-to-end latency per decision, dominated by the HTTP request time, well within the bounds required for synchronous action interception. \sysname{} works synergistically with the
cryptographic credential infrastructure of frameworks such as A2AS\@.
The key design insight is the separation of two concerns that flat-list
engines conflate: \emph{what} an agent is allowed to do (permission),
and \emph{what follows} from allowing it (obligation).  Deontic logic
makes this separation explicit. Enforcing it deterministically, at
scale, in production agentic systems is the challenge we are working
toward.

\section*{Acknowledgments}
This work was partially funded by NSF award 2310844, IUCRC Phase II UMBC: Center for Accelerated Real-Time Analytics (CARTA). Generative AI systems were used to create the structure of the document, curate related work, check grammar, and refine the presentation. They were also used to help code the policy enforcement prototype.

\bibliographystyle{IEEEtran}
\bibliography{refs}

\end{document}